\documentclass[letterpaper, 10 pt, conference]{ieeeconf}  

\IEEEoverridecommandlockouts                              

\usepackage[style=numeric,sorting=none,maxbibnames=6]{biblatex}

\usepackage{colortbl}
\usepackage{url, amsfonts, amsmath, hyperref, subcaption}
\usepackage{dblfloatfix}    
\usepackage{graphicx}

\usepackage{gensymb}
\usepackage{multirow}
\usepackage{MnSymbol}
\usepackage{wasysym}
\usepackage{float}
\usepackage[x11names]{xcolor}
\usepackage{soul}
\usepackage[normalem]{ulem}

\usepackage[labelsep=period,skip=1pt]{caption}
\usepackage{dblfloatfix} 
\usepackage{etoolbox}
\newtoggle{vtwo}
\newtoggle{vtwo_hl}
\newtoggle{vfinal}
\newtoggle{vfinal_hl}

\togglefalse{vtwo}
\togglefalse{vtwo_hl}
\togglefalse{vfinal}
\togglefalse{vfinal_hl}

\usepackage{array, booktabs, makecell} 
\usepackage{siunitx, mhchem}

\usepackage{multirow}
\usepackage{caption}

\usepackage{bm}

\title{

\LARGE \bf An Untethered Bioinspired Robotic Tensegrity Dolphin with Multi-Flexibility Design for Aquatic Locomotion}

\author{Albert Author$^{1}$ and Bernard D. Researcher$^{2}$
\thanks{$^{1}$Albert Author is with Faculty of Electrical Engineering, Mathematics and Computer Science,
        University of Twente, 7500 AE Enschede, The Netherlands
        {\tt\small albert.author@papercept.net}}%
\thanks{$^{2}$Bernard D. Researcheris with the Department of Electrical Engineering, Wright State University,
        Dayton, OH 45435, USA
        {\tt\small b.d.researcher@ieee.org}}%
}
\author{Luyang Zhao$^{1,*}$, Yitao Jiang$^{1,*}$, Chun-Yi She$^{1}$, Mingi Jeong$^{1}$, \\ Haibo Dong$^{2}$, Alberto Quattrini Li$^{1}$, Muhao Chen$^{3}$, Devin Balkcom$^{1}$%
\thanks{$^{*}$ means contributed equally to this work.}%
\thanks{$^{1}$Luyang Zhao, Yitao Jiang, Chun-Yi She, Mingi Jeong, Alberto Quattrini Li, and Devin Balkcom are with the Department of Computer Science, Dartmouth College, Hanover, NH, USA
        {\tt\small \{luyang.zhao.gr, yitao.jiang.gr, chun-yi.she.gr, mingi.jeong.gr, alberto.quattrini.li, devin.balkcom\}@dartmouth.edu}}%
\thanks{$^{2}$Haibo Dong is with the Department of Mechanical and Aerospace Engineering, University of Virginia, Charlottesville, VA, USA
        {\tt\small hd6q@virginia.edu}}%
\thanks{$^{3}$Muhao Chen is with the Department of Computer Science, University of Kentucky, Lexington, KY, USA
        {\tt\small muhaochen@uky.edu}}%
}

\begin{document}

\maketitle

\begin{abstract}
 
This paper presents the first steps toward a soft dolphin robot using a bio-inspired approach to mimic dolphin flexibility. The current dolphin robot uses a minimalist approach, with only two actuated cable-driven degrees of freedom actuated by a pair of motors. The actuated tail moves up and down in a swimming motion, but this first proof of concept does not permit controlled turns of the robot. While existing robotic dolphins typically use revolute joints to articulate rigid bodies, our design -- which will be made opensource -- incorporates a flexible tail with tunable silicone skin and actuation flexibility via a cable-driven system, which mimics muscle dynamics and design flexibility with a tunable skeleton structure. The design is also tunable since the backbone can be easily printed in various geometries. The paper provides insights into how a few such variations affect robot motion and efficiency, measured by speed and cost of transport (COT). This approach demonstrates the potential of achieving dolphin-like motion through enhanced flexibility in bio-inspired robotics.

\end{abstract}

\section{Introduction}
\label{sec:introduction}

Dolphins swimming energy efficiency has long intrigued researchers and engineers~\cite{fish1991dolphin,leatherwood2012bottlenose,shane1986ecology}. This efficiency is largely attributed to their whole-body flexibility, enabling dynamic shape adaptation to support diverse swimming maneuvers \cite{guo2023thrust, Han020forward, wang21computational}. The study of dolphin locomotion has provided valuable insights into the design of aquatic robots, particularly for applications that require energy-efficient underwater mobility~\cite{PRADEEP2023120941}.

Recent advances in bio-inspired robotics have sought to replicate these capabilities in artificial systems. Most robotic dolphins still rely on traditional mechanical systems, such as revolute joints and rigid structures, which, while enabling the robot to swim, often fail to capture the natural fluidity of a dolphin’s movement \cite{Chen2023performance, Tanaka2019burst, yu2008dolphin, yu2016development, wu2019towards}. The limitations of rigid designs may restrict the robot's ability to perform the smooth, continuous deformations essential for efficient swimming. Tendon-driven designs inspired by dolphin anatomy have been barely explored. 
One particular dolphin robot by Liu et al. has shown promising results in forward swimming speed \cite{liu2021design}; however, its fixed vertebra design limits the exploration of alternative skeletal configurations.

\begin{figure}[tp]
    \centering
    \includegraphics[width=\linewidth]{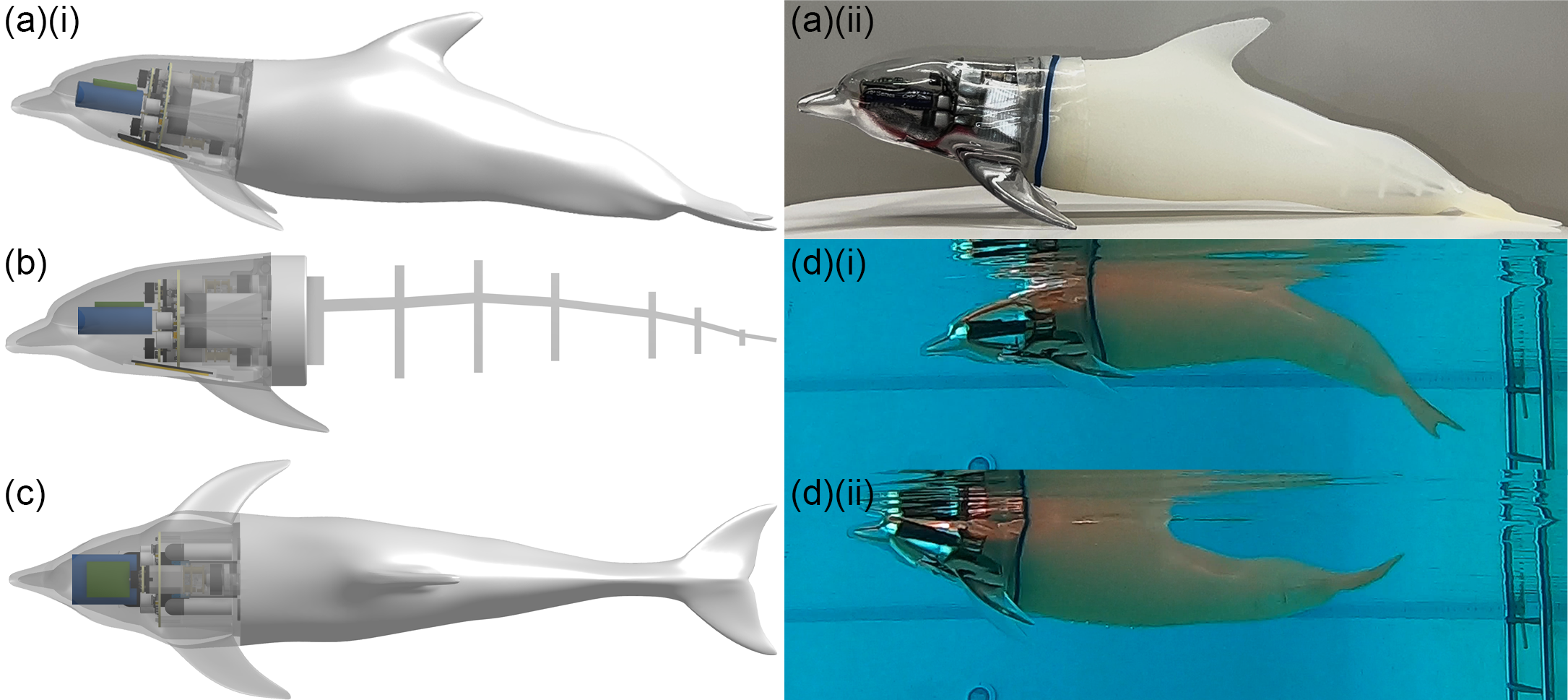}
    \caption{Overview of the dolphin robot: (a)(i) Side view of CAD model, (a)(ii) Side view of the real robot, (b) Dolphin robot showing inner skeleton, (c) Top view of CAD model, (d)(i) Swimming with tail down, (d)(ii) Side view of the real dolphin robot swimming with tail up.}
    \label{fig:overview}
\end{figure}

Soft robotics has emerged as a promising field for replicating the flexibility found in nature~\cite{zou2018reconfigurable, nemitz2016using,softsnap}. These robots, made of compliant materials like silicone and elastomers~\cite{buckner2021design, ShepherdRobertF2011Msr, Tolley2014}, exhibit adaptable, fluid motion that rigid systems cannot achieve~\cite{lee2017soft}. Soft robots can deform in response to environmental forces, enabling them to navigate complex terrains and perform precise, delicate movements~\cite{softlattice, zebing}. 

By incorporating flexible materials, a bio-inspired robotic dolphin can better emulate the dynamic properties of a real dolphin's body, enhancing flexibility while maintaining reasonable energy efficiency. While previous dolphin robots have predominantly utilized rigid materials with revolute or other joint mechanisms, our design is the first to employ silicone-based, compliant materials for the dolphin's body, allowing smoother, more lifelike aquatic motions. Additionally, flexible skeletons driven by cable or shape memory alloy (SMA) systems~\cite{Lai2022cable, shintake2020fish, starblocks} and designs like the fishbone-inspired framework~\cite{softsnap} further contribute to this fluid movement.

To address the challenges of replicating dolphin motion, we introduce the first robotic dolphin design to integrate flexibility across three key aspects: body, actuation, and structural adaptability. Body flexibility (Section~\ref{sec:body-flex}) is achieved by pairing a rigid head with a flexible tail, covered in a tunable silicone skin that allows hardness adjustments. This unique feature enables the robot to mimic the adaptive properties of real dolphins, enhancing movement efficiency. Actuation flexibility (Section~\ref{sec:actuation-flex}) is realized through a cable-driven system that emulates muscle contractions, creating smooth, continuous motion that closely resembles dolphin muscle dynamics. A fishbone mechanism within the actuation system further enhances motion adaptability and fluidity. Lastly, design flexibility (Section~\ref{sec:design-flex}) is incorporated through a compliant skeleton structure, which design can be tuned in terms of pin bones length ratio, enabling a family of designs for exploring parameters that influence swimming speed and efficiency. The design will be made opensource.

This paper presents the detailed design and development of a bio-inspired robotic dolphin, examining how each aspect of flexibility contributes to achieving dolphin-like motion. We discuss the geometric modeling of the skeleton, robot fabrication, and results from swimming tests that evaluate the design's performance (Section~\ref{sec:test}). The current prototype allows the robot to swim only forward, but this work is a first step aiming to advance bio-inspired robotics by demonstrating how flexible design can enhance the locomotion of underwater robots -- Section~\ref{sec:conclude} discusses corresponding future work. 

\section{Body Flexibility}
\label{sec:body-flex}
\subsection{Bottlenose Dolphin Model}

The design of our robotic dolphin emphasizes replicating the natural flexibility required for efficient swimming. A flexible tail mimics the undulating movements of real dolphins, while the electronic components are securely housed within a rigid head for protection. The body is divided into two sections: the head, which comprises approximately 33\% of the total length, and the tail, making up the remaining 67\%. This division balances the need for flexibility with the safeguarding of essential components.

\begin{figure}[htbp]
    \center
    \includegraphics[width=.7\linewidth]{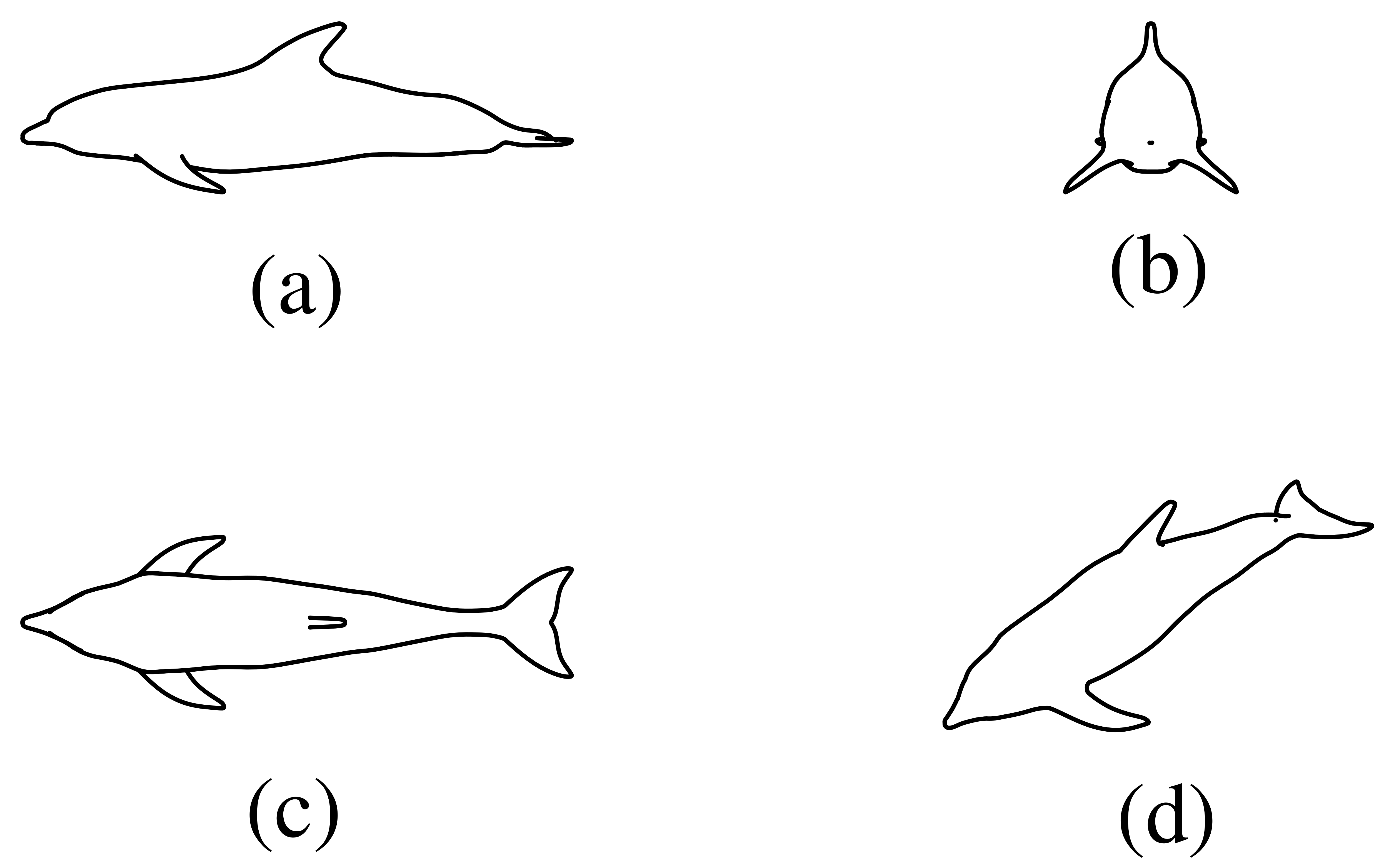}
    \caption{3D Dolphin Model: (a) Side view, (b) Front view, (c) Top view, and (d) Oblique view.}
    \label{fig:whole_sys}
\end{figure}

A bottlenose dolphin was selected as the biological model for this design, given the extensive research available on its swimming mechanics~\cite{guo2023thrust}. The geometric model of the dolphin is shown in Fig.~\ref{fig:whole_sys}.

\subsection{Flexible Tail}

\begin{figure}[htbp]
    \centering
    \includegraphics[width=\linewidth]{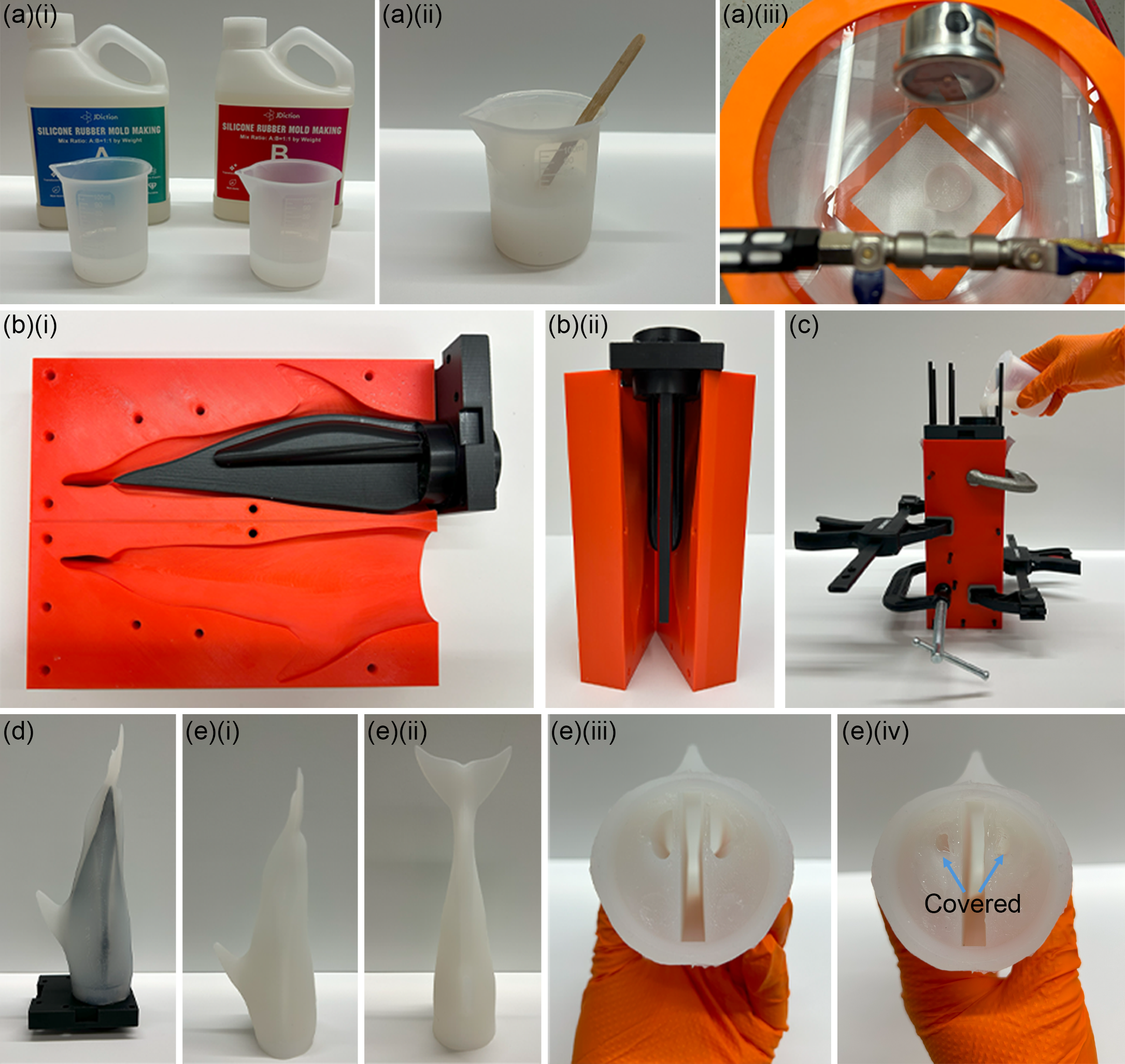}
    \caption{Silicone Tail Molding and Demolding Process: (a) Molding steps: (i) Measuring silicone solutions, (ii) Stirring the mixture, (iii) Vacuum chamber treatment; (b) Mold preparation: (i) 3D-printed mold components, (ii) Pre-assembled mold; (c) Pouring silicone into the mold. Demolding the Silicone Tail: (d) Tail during demolding; (e) Final views of the tail: (i) Left view, (ii) Front view, (iii) Side view showing the two air chambers, (iv) Sealing the air chambers for waterproofing and buoyancy.}
    \label{fig:mold_demold}
\end{figure}

The tail is crafted from molded silicone, specifically with 10A hardness, chosen for its flexibility and ability to return to its original shape, effectively emulating a dolphin's natural swimming motion. 
As shown in Fig.~\ref{fig:mold_demold}, the fabrication process involves multiple steps: silicone solutions A and B are first measured and mixed, following the manufacturer instructions, at a 1:1 volume ratio; stirred for 5 minutes; and then degassed in a vacuum chamber for 15 minutes to remove air bubbles. 
Next, a clamp is used to secure the 3D-printed mold, into which the silicone mixture is carefully poured. After curing for over 8 hours, the tail is removed from the mold. The process has proven to be highly repeatable, allowing us to fabricate multiple tails with consistent characteristics. The cured tail includes two internal air chambers designed to provide slight positive buoyancy, allowing the robot to float vertically in water. To ensure waterproofing, we sealed the chambers using a membrane and liquid silicone glue.

\begin{figure}[htbp]
    \centering
    \includegraphics[width=1.1\linewidth,trim={2cm 0 0 0},clip]{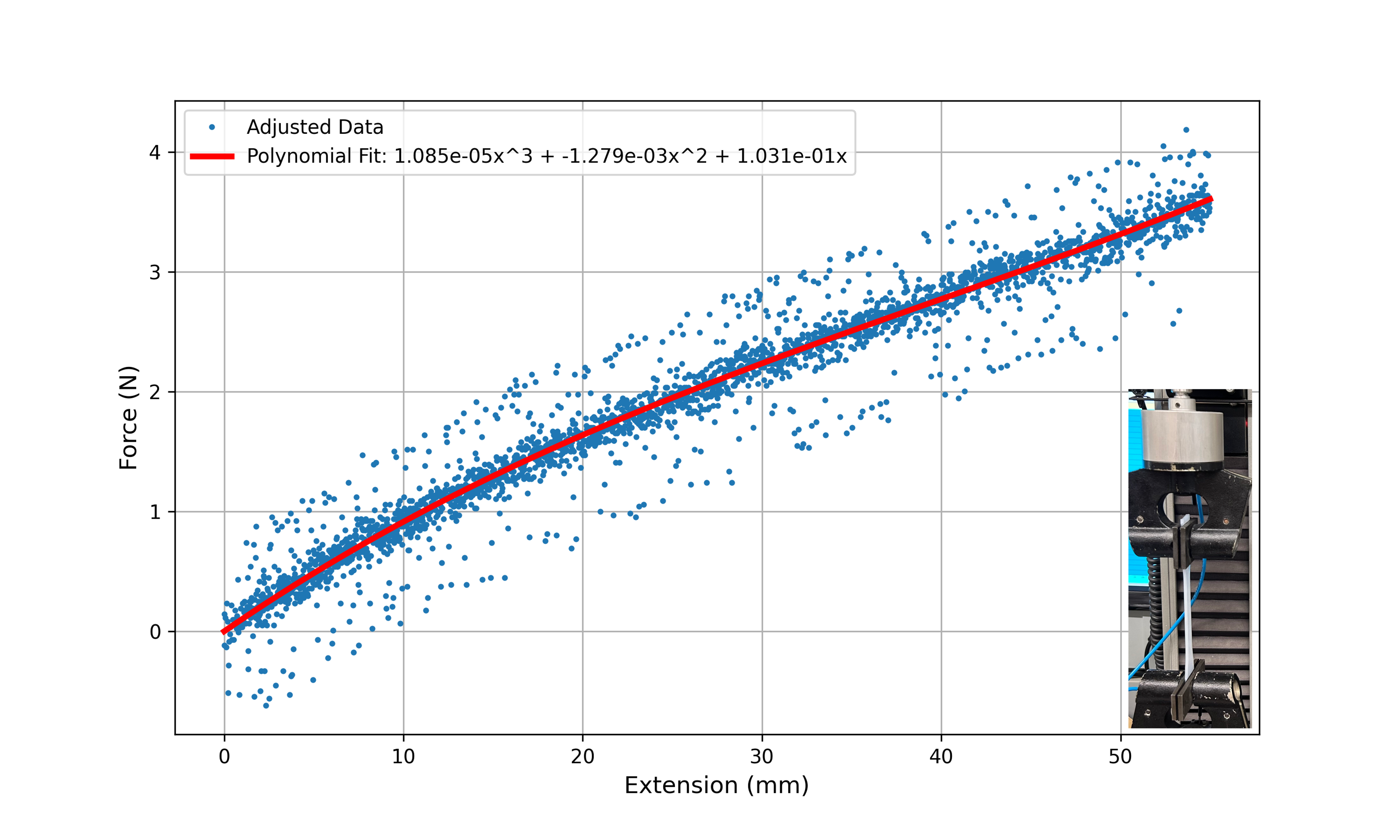}
    \caption{Force-displacement analysis of the silicone used for the dolphin’s tail, conducted using an Instron machine.}
    \label{fig:force-displace}
\end{figure}

Mechanical testing was conducted on an Instron machine to assess the silicone’s force-displacement properties, as illustrated in Fig.~\ref{fig:force-displace}. The results provided insights into the tail’s stiffness.

\subsection{Rigid Head}

\begin{figure}[htbp]
    \centering
    \includegraphics[width=\linewidth]{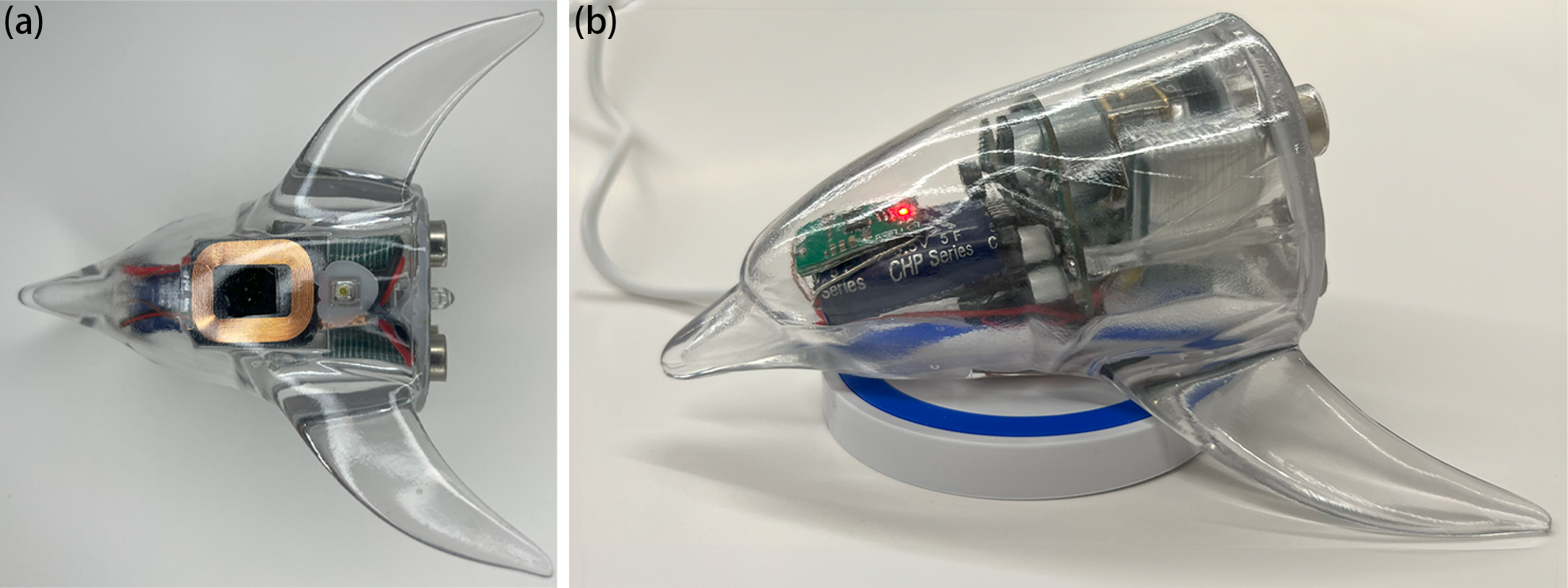}
\caption{Rigid head of our robotic dolphin: (a) Bottom view showing the wireless charging coil, and (b) The head charging on a Qi wireless charger, with the red indicator light illuminated.}
    \label{fig:head}
\end{figure}

\begin{figure}[htbp]
    \centering
    \includegraphics[width=\linewidth, trim=0mm 0mm 0mm 0mm, clip]{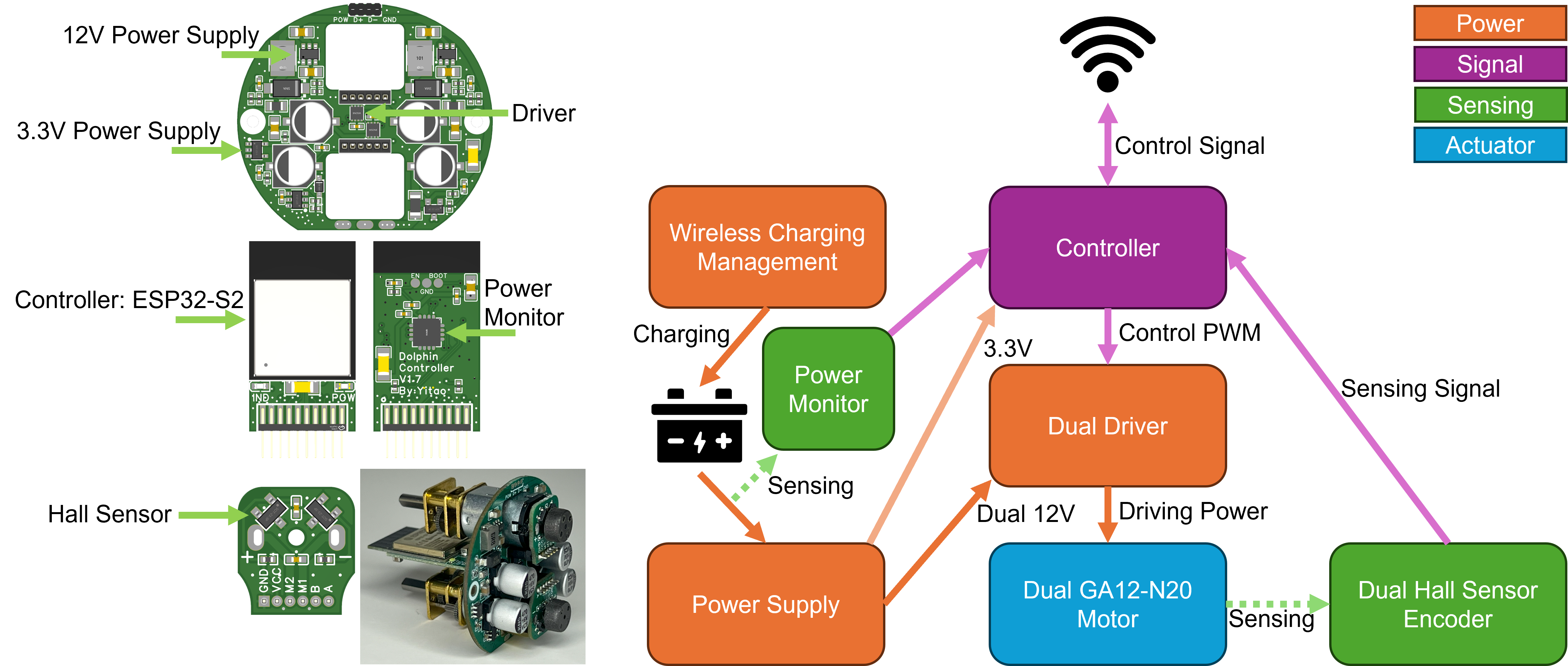}
        \caption{Detailed PCB design and functionality flow chart overview, illustrating the design concept, which includes components for power distribution, signal processing, sensing, and actuation. The diagram highlights the integration of power supply units, the controller (ESP32-S2), drivers, sensors, and wireless charging management.}
    \label{fig:pcb}
\end{figure}

The rigid head, shown in Fig.~\ref{fig:head}, houses all essential electronic components, including custom-designed printed circuit boards (PCB) for power distribution, signal processing, sensing, and actuation (details in Fig.~\ref{fig:pcb}), two 702025 lithium-ion batteries (3.7 V, 250 mAh, 20C), a 5.5V 5F supercapacitor, and Wi-Fi communication for control. This compact arrangement, protected from mechanical stress and water exposure, provides efficient power management: the high-density batteries support sustained operation with minimal bulk, while the supercapacitor delivers instantaneous high current for motor actuation. Concentrating the electronics in the head enhances stability, allowing the flexible tail to move freely and mimic the swimming mechanics of a dolphin.

\begin{figure}[htbp]
    \centering
    \includegraphics[width=\linewidth, trim=0mm 0mm 0mm 0mm]{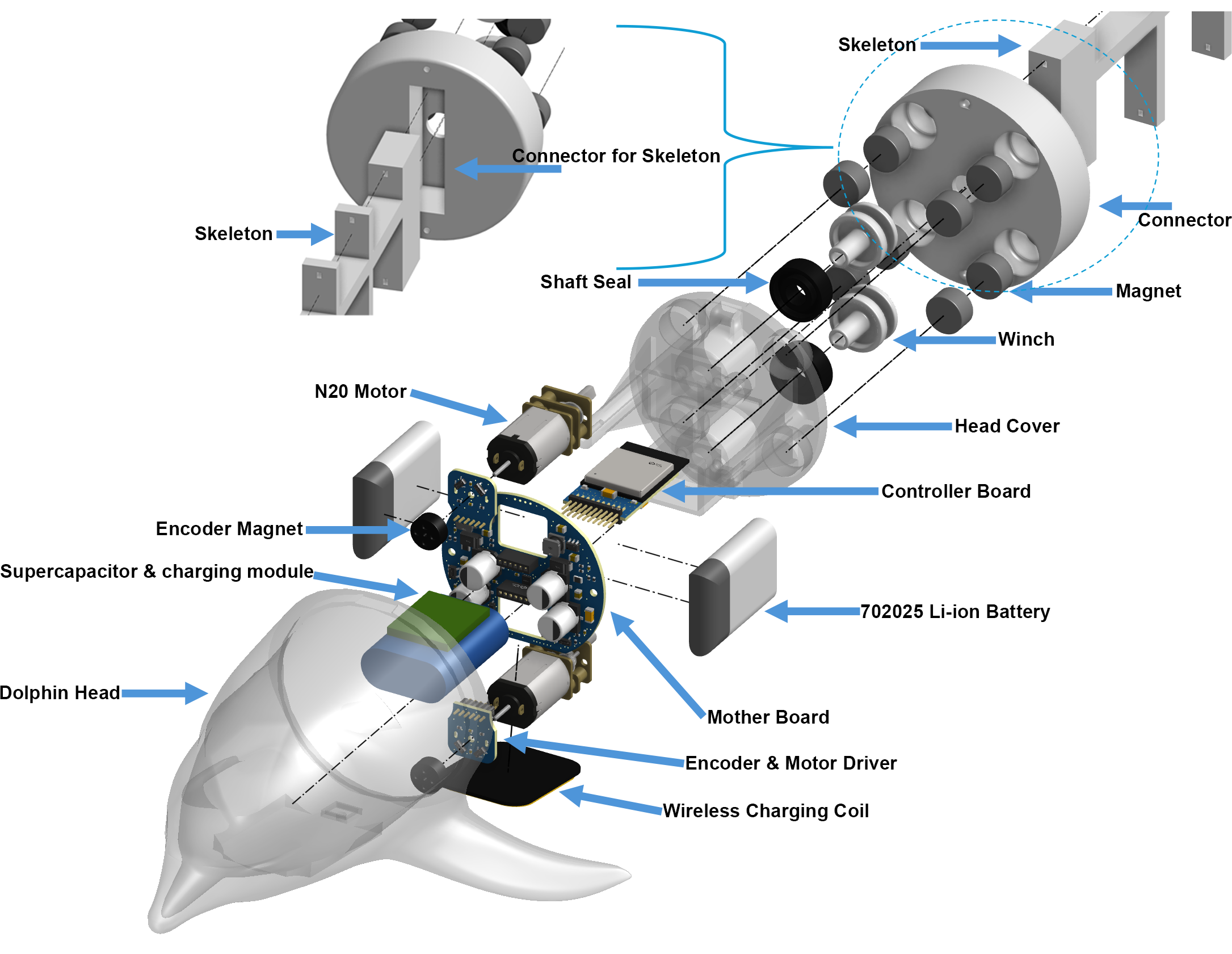}
    \caption{Exploded view of the robotic dolphin's head and connectors, highlighting the arrangement of electrical components within the rigid head and the magnets embedded in the two connectors.}
    \label{fig:exploded-view}
\end{figure}

The head and skeleton are connected with two connectors containing permanent magnets, enabling easy tail replacement with other designs (Fig.~\ref{fig:exploded-view}). The silicone tail attaches to the skeleton, secured with removable waterproof silicone tape that forms a watertight seal through simple contact, allowing for effective waterproofing and easy detachment.

\section{Actuation Flexibility}
\label{sec:actuation-flex}
To replicate the complex muscle dynamics of a dolphin, we developed a flexible actuation system using a combination of a cable-driven mechanism and a fish-bone skeleton structure. This system enables smooth, adaptive movements, closely mimicking the natural swimming patterns of a dolphin. The fish-bone skeleton provides the structural framework, while the cable-driven system simulates muscle contractions, allowing for dynamic and efficient propulsion.

\subsection{Fish-Bone Skeleton Design}

\begin{figure}[htbp]
    \center
    \includegraphics[trim=0cm 0cm 0cm 0cm, clip=false,width=\linewidth]{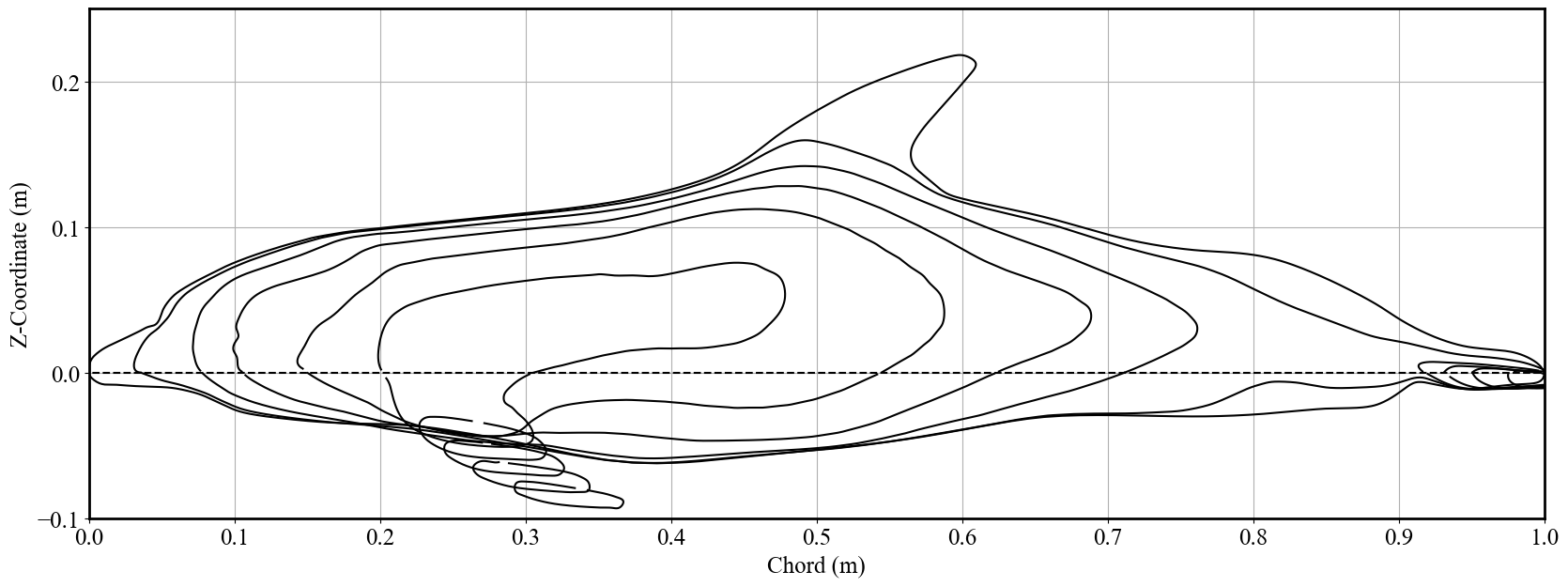}
    \caption{Hull lines of the dolphin model along the Y-axis. Cut in 10 slides in half of the body with equal spacing. End-cut excluded.}
\label{fig:hulline-y}
\end{figure}

\begin{figure}[htbp]
    \center
    \includegraphics[trim=0cm 0cm 0cm 0cm, clip=false,width=\linewidth]{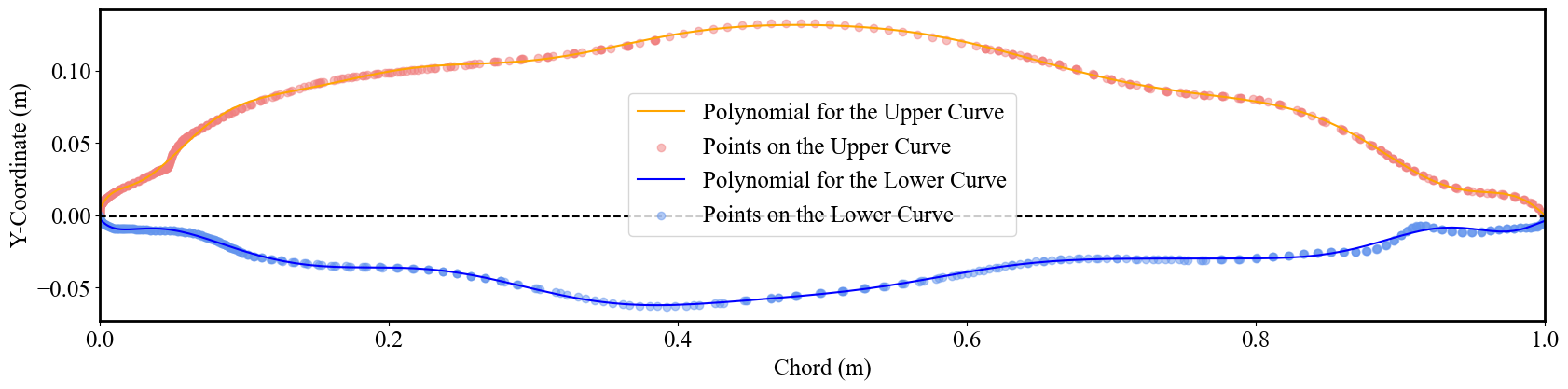}
    \caption{Discretized center cut dolphin body.}
\label{fig:2d_points}
\end{figure}

The fish-bone skeleton design is inspired by the segmented vertebrae of real dolphins and serves as the core structural framework of the robotic dolphin. This skeleton is derived from sequential cross-sectional cuts along the dolphin’s body length (Fig.~\ref{fig:hulline-y}), with the dorsal fin removed from this design phase to streamline the profile. To achieve this, data points from 0.40 m to 0.61 m along the x-axis were excluded, creating a simplified dorsal profile for improved suitability in robotic applications. The resulting data points define the basic geometry of the skeleton, as shown in Fig.~\ref{fig:2d_points}.

To create a continuous upper curve, interpolation fills gaps left by the removed fin. Specifically, cubic interpolation (from `scipy.interp1d' function) generates additional data points along the dorsal contour, ensuring a seamless, uninterrupted shape. 

For precise mathematical representation, polynomial curve fitting is applied to both the upper and lower curves using the following 17th-order polynomials:
\begin{align}
        y_u = \sum_{i=0}^{17}a_{i} x^{i}, \quad y_l = \sum_{i=0}^{17}b_{i} x^{i},
\end{align}
where \( x \) represents scaled coordinates along the chord (from 0 to 1 m), and \( \bm{a} \) and \( \bm{b} \) are coefficients computed with a least square polynomial fitting (`numpy.polyfit`). This fitting process yields low mean squared errors of \( 2.4514 \times 10^{-6} \, \text{m}^2 \) and \( 1.4484 \times 10^{-6} \, \text{m}^2 \), ensuring an accurate depiction of the dolphin’s streamlined body.

\begin{figure}[htbp]
    \centering
    \includegraphics[width=\linewidth,trim={0.8cm 0 1cm 0},clip]{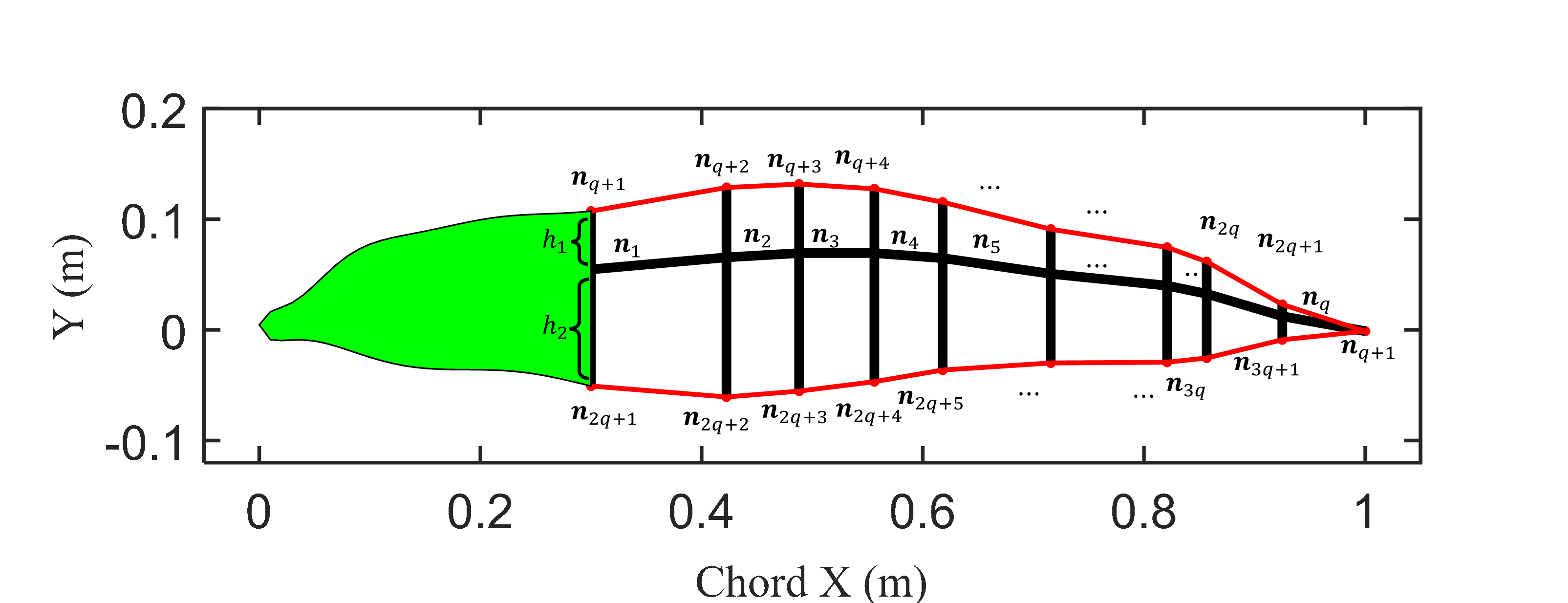}
    \caption{Fish-bone skeleton configuration of our tensegrity dolphin: the green area represents the rigid body head, which comprises approximately 33\% of the total length, while the black and red lines represent bars and strings, respectively. }
    \label{notation_nodes_without_X}
\end{figure}

The skeleton design uses a tensegrity-inspired structure composed of rigid bars and flexible tendons (shown in Fig.~\ref{notation_nodes_without_X}) to emulate the dolphin’s flexible spine. This segmented arrangement enables controlled tail flexion, balancing flexibility, and stability for efficient swimming. The fish-bone skeleton distributes forces evenly across the body, preventing strain concentration and enabling smooth, natural movements even during complex maneuvers.

This compact and lightweight 2D skeleton is ideal for aquatic environments where maneuverability is critical. It provides a strong yet flexible foundation that supports the cable-driven system, allowing the robotic dolphin to adapt its movements in response to external conditions.

\subsection{Cable-Driven System}

The cable-driven system forms the core of the dolphin’s actuation, simulating muscle contractions by controlling the length of two cables that run along the fish-bone skeleton. These cables are anchored at key points along the skeleton and are adjusted by GA12-N20 motors with a 1:100 gear ratio located in the rigid head of the robot. This motor, capable of reaching 50,000 RPM, provides a balanced output of torque and speed. Encoders on each motor allow precise control over the length of each cable, enabling the system to achieve the desired tail curvature and flexion for various swimming motions. 

By controlling the cable length, the system can produce smooth, wave-like motions essential for effective propulsion. This length control allows the tail to curve and flex in response to environmental changes, enabling adaptive swimming patterns. For example, when greater thrust is needed, the cable length can be shortened to create a more pronounced wave, increasing the speed of the swimming motion. Conversely, the cable length can be increased for gentler maneuvers.

The fish-bone skeleton is 3D-printed using PLA material, offering a lightweight yet durable framework. The use of lightweight cables and PLA construction reduces the system's overall weight, enhancing the robot’s agility and responsiveness in water. Additionally, placing the motors in the rigid head centralizes the actuation system, ensuring that the tail remains free to flex and bend without interference.

The power consumption of the system is about 0.48W in idle and 9.33W in Type 4 actuation. The battery provides 1.85Wh power. With such a small battery (for comparison, a smartphone can have a 20Wh battery), the running time in idle is 3.85h and 0.2h with Type 4 actuation.

\subsection{Integrated Actuation and Flexibility}

By integrating the fish-bone skeleton with the cable-driven system, we achieved a balance between flexibility and structural integrity. The skeleton provides the necessary support to maintain the shape of the tail during high-stress maneuvers, while the cable-driven system enables the dynamic, adaptive motions required for swimming. This combination allows the robotic dolphin to replicate the efficient, undulating motions of a real dolphin, adapting seamlessly to changes in the aquatic environment.

The system’s design also enhances the overall efficiency of the robot by minimizing energy consumption during swimming. The use of two actuating cables simplifies the control mechanism, while the skeleton’s lightweight nature reduces the energy required to move the tail through the water. This ensures that the robot can sustain prolonged operation while maintaining optimal performance.

\section{Design Flexibility}
\label{sec:design-flex}
The skeleton of the robotic dolphin is designed to be highly tunable, allowing a range of design variations by adjusting parameters such as the number of ribs, rib thickness, the shape of the middle rod, the height ratio \( h_1:h_2 \) (Fig.~\ref{notation_nodes_without_X}), and the thickness ratio. This tunability aims to optimize both speed and COT, enabling a detailed exploration of the trade-offs between these performance metrics. By examining different configurations, we can identify a Pareto front of solutions, balancing efficient locomotion with energy consumption to achieve optimal performance.

\begin{figure}[htbp]
    \centering
    \includegraphics[width=\linewidth]{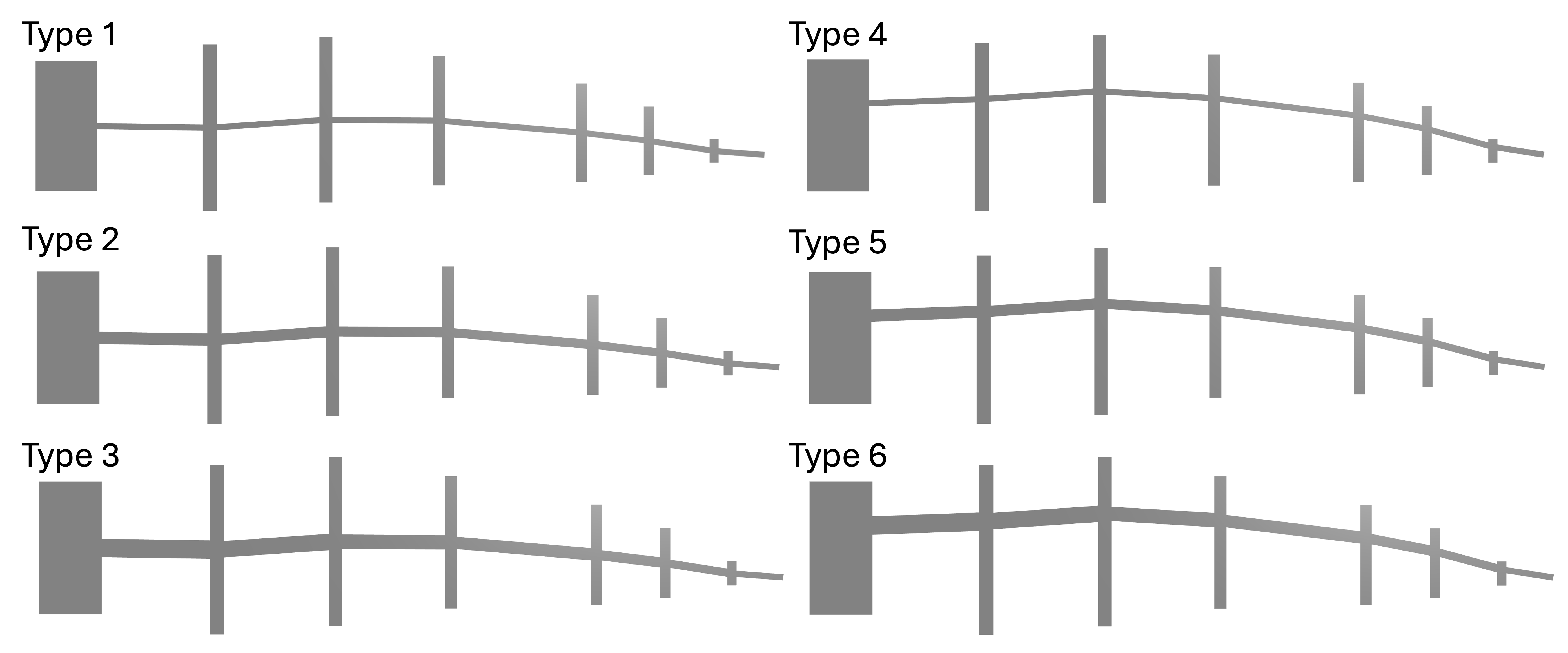}
    \caption{Six types of skeletons. The specifications of each skeleton are shown in Table~\ref{tab:robot-speed-cot}.}
    \label{fig:6-skeleton}
\end{figure}

The thickness ratio refers to the ratio between the thickness of the first rib (the one closest to the head) and the final rib, with intermediate rib thicknesses interpolated accordingly. For the spine sections between two ribs, the thickness matches that of the adjacent rib closer to the head. In this study, we focus on two key parameters: the \( h_1:h_2 \) ratio, where a smaller ratio results in higher tail and the rib thickness ratio, where a higher thickness, requires more force for bending and also influences which segment of the tail will be bended. Six examples of skeleton designs are shown in Fig.~\ref{fig:6-skeleton}, and Table~\ref{tab:robot-speed-cot} summarizes these parameters and their impact on tail movement and energy efficiency.

Additionally, we designed a modular connector system (Fig.~\ref{fig:exploded-view}) for easy replacement of skeleton types, streamlining the process for testing various configurations. This modular approach simplifies manufacturing and allows rapid reconfiguration of the robot, facilitating the efficient exploration of optimal design solutions.

\section{Swimming Test}
\label{sec:test}
\subsection{Test Setup}

We 3D-printed six different types of skeletons, systematically going through the combinations of two $h_1: h_2$ ratios and three thickness ratios. Each skeleton was assembled with the same silicone dolphin tail to test their swimming performance in a swimming pool, which was recorded by placing a GoPro camera underwater to capture their movements.

Swimming performance (results in Table~\ref{tab:robot-speed-cot})  is measured with three metrics: speed \( v \) in mm/s, speed in body length per second (bl/s), and COT.
The swimming speed was obtained by analyzing the recorded GoPro footage and tracking the robot's displacement over a fixed distance. The COT for each configuration was calculated using the following formula:
\begin{equation}
\text{COT} = \frac{P}{m v},
\end{equation}
where \( P \) is the average power consumption of the robot during swimming (in watts), \( m \) is the mass of the robot (in kilograms), \( v \) is the average swimming speed (in meters per second).
To measure \( P \), we used a current and voltage sensor connected to the robot’s motor to record the power draw over time. 

The COT value allows us to compare the energy efficiency of each skeleton configuration by normalizing power consumption with respect to both speed and mass, providing a reliable metric for assessing swimming performance across different skeleton types.

\begin{table}[htbp]
    \centering
    \caption{Robot Speed and COT for Different Skeleton Types.}
    \resizebox{\columnwidth}{!}{%
    \begin{tabular}{|c|c|c|c|c|c|}
        \hline
        \textbf{Skeleton Type} & \textbf{$h_1: h_2$} & \textbf{Thickness Ratio} & \textbf{Speed (mm/s)} & \textbf{Speed (bl/s)} & \textbf{COT} \\
        \hline
        Type 1 & 1:1 & 1:1 & 133.5607 & 0.411 & 146 \\ 
        \hline
        Type 2 & 1:1 & 2:1 & 125.4027 & 0.386 & 136 \\ 
        \hline
        Type 3 & 1:1 & 3:1 & 127.8671 & 0.393 & 136 \\ 
        \hline
        \rowcolor{lightgray} 
        Type 4 & 1:2 & 1:1 & 163.1813 & 0.502 & 95 \\ 
        \hline
        Type 5 & 1:2 & 2:1 & 86.8601 & 0.267 & 175 \\
        \hline
        Type 6 & 1:2 & 3:1 & 78.7879 & 0.243 & 193 \\ 
        \hline
    \end{tabular}
    }
    \label{tab:robot-speed-cot}
\end{table}

\begin{figure}[htbp]
    \centering
    \includegraphics[width=\linewidth]{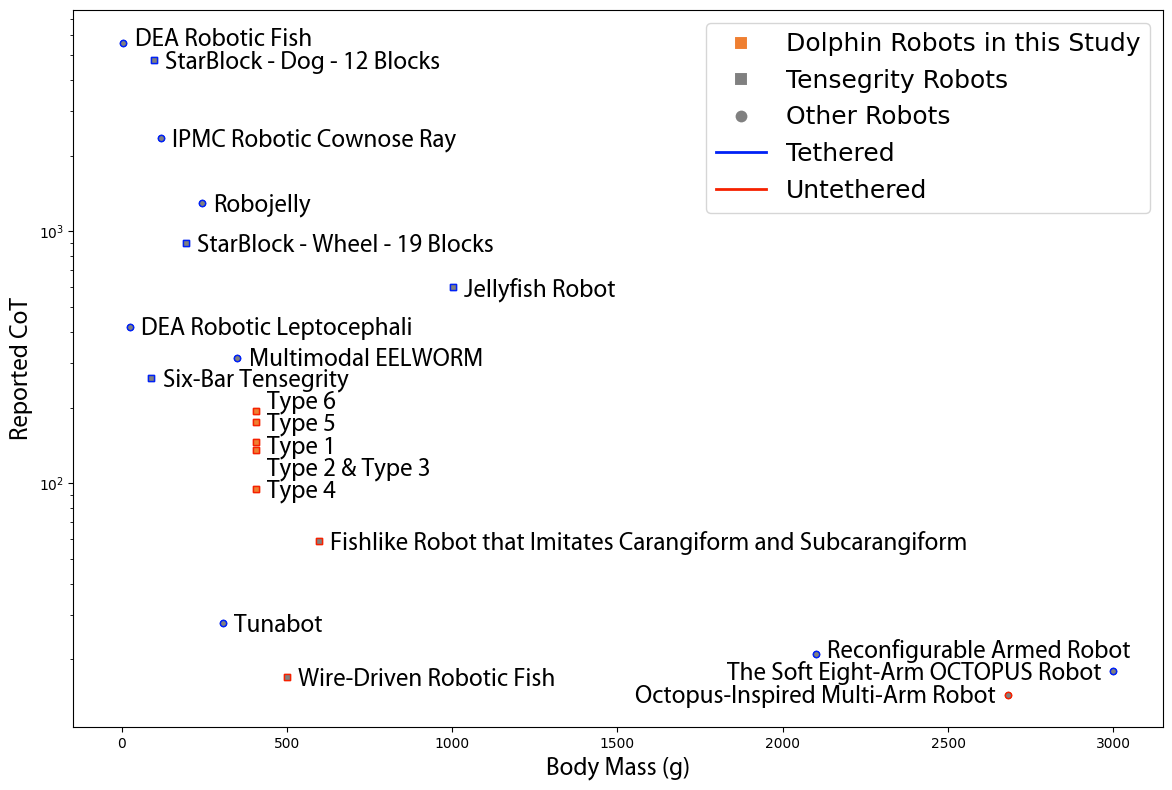}
    \caption{The COT comparison among six skeleton robot designs and other robotic systems. All referenced data can be found in \cite{starblocks,White_2021,Rieffel}.}
    \label{fig:COT-comparison}
\end{figure}

\begin{figure*}[htbp]
    \centering
    \includegraphics[width=\linewidth, trim=0mm 0mm 0mm 0mm, clip]{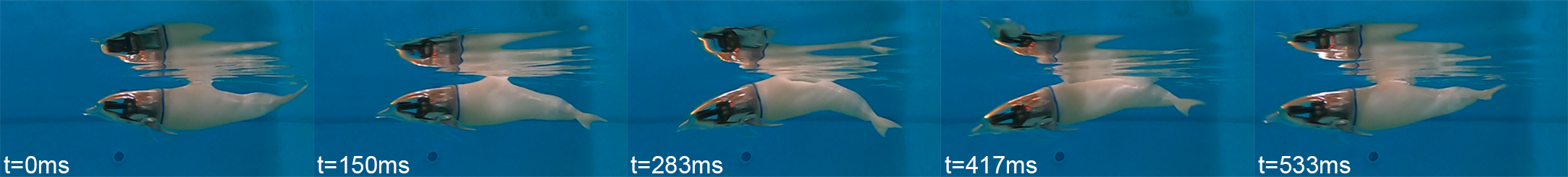}
    \caption{Sequence of motion screenshots for Skeleton Type 4 during swimming.}

    \label{fig:motion-sequence}
\end{figure*}

\subsection{Results and Analysis}

The results in Table~\ref{tab:robot-speed-cot} show that Skeleton Type 4 achieves the highest speed at 163.18 mm/s while also yielding the lowest COT at 95, indicating optimal energy efficiency among the tested configurations. A sequence of motion screenshots illustrating this skeleton type’s swimming performance is shown in Fig.~\ref{fig:motion-sequence}.
This skeleton type has a ratio of \( h_1:h_2 = 1:2 \) and a thickness ratio of 1:1, suggesting that a higher \( h_1:h_2 \) ratio combined with a balanced thickness offers advantages in both speed and energy efficiency. 

Skeleton Types 1, 2, and 3, which have an \( h_1:h_2 \) ratio of 1:1 (lower tail height), display moderate speeds ranging from 125.40 mm/s to 133.56 mm/s, with COT values around 136. These configurations provide a balance between speed and energy consumption but are outperformed by Skeleton Type 4 in terms of both metrics. 

On the other hand, Skeleton Types 5 and 6, which also use a ratio of \( h_1:h_2 = 1:2 \) but with larger thickness ratios of 2:1 and 3:1, respectively, exhibit the lowest swimming speeds and the highest COT values (175 and 193). This indicates that increased thickness negatively impacts both speed and energy efficiency. 

As shown in Fig~\ref{fig:COT-comparison}, our module shows lower efficiency compared to some aquatic robots, such as the tunabot and fish-like robot~\cite{White_2021}, but it performs more efficiently than others like the robojelly and DEA robotic fish~\cite{White_2021}. Moreover, many aquatic robots are tethered, including various types of fish-inspired and multi-arm designs, while a smaller set, like our dolphin robot, the wire-driven robotic fish, and the octopus-inspired multi-arm robot, operate untethered.

\section{Conclusion and Future Work}
\label{sec:conclude}

This study introduces a new untethered tensegrity-inspired dolphin robot with multi-flexibility features. We evaluated the swimming performance of this modular robotic platform using various 3D-printed skeleton configurations to identify the optimal design for energy-efficient underwater locomotion. Among six skeleton types, Skeleton Type 4—with an \( h_1:h_2 \) ratio of 1:2 and a balanced thickness of 1:1—achieved the highest speed and lowest COT, indicating superior energy efficiency, thereby enhancing swimming performance.

These findings emphasize the potential of customized skeleton designs for energy-efficient robotic swimming. Selecting configurations with suitable height and thickness ratios can optimize performance for underwater exploration, marine research, and other aquatic applications.

Future work will further refine structural parameters, test additional ratios, and explore adaptive skeleton designs to improve versatility in diverse underwater environments. Additionally, we aim to integrate controlled turning and enhanced maneuverability, advancing the development of energy-efficient underwater robotic systems. To enable autonomous operation, we plan to incorporate a camera into the robot's head for real-time visual feedback and environmental awareness. Furthermore, we will add steering mechanisms for precise navigation and buoyancy control systems to enhance stability and depth regulation, enabling more complex and efficient underwater behaviors.

\printbibliography
\end{document}